\documentclass[10pt]{article}
\usepackage[accepted]{tmlr}
\usepackage{amsmath,amsfonts,bm}

\def\eqref#1{equation~\ref{#1}}
\def\1{\bm{1}}

\DeclareMathAlphabet{\mathsfit}{\encodingdefault}{\sfdefault}{m}{sl}
\SetMathAlphabet{\mathsfit}{bold}{\encodingdefault}{\sfdefault}{bx}{n}

\usepackage{hyperref}
\hypersetup{pdftitle={Causal Scene Narration: Inference-Time Text Restructuring for Vision-Language-Action Driving}, pdfauthor={Yun Li, Yidu Zhang, Simon Thompson, Ehsan Javanmardi, Manabu Tsukada}}
\usepackage{url}
\usepackage{amsmath,amsfonts,amssymb}
\usepackage{graphicx}
\usepackage{booktabs}
\usepackage{placeins}
\usepackage{bm}
\usepackage{xcolor}
\definecolor{revBlue}{rgb}{0.0, 0.3, 0.7}
\definecolor{revOrange}{rgb}{0.85, 0.45, 0.0}
\definecolor{revTeal}{rgb}{0.0, 0.5, 0.45}
\newcommand{\rOne}[1]{#1}    % Reviewer uZ8t (#1)
\newcommand{\rTwo}[1]{#1}  % Reviewer 2g4t (#2)
\newcommand{\rThree}[1]{#1}  % Reviewer Lwcg (#3)
\usepackage{multirow}
\usepackage{float}
\usepackage{pifont}
\usepackage[normalem]{ulem}
\usepackage{algorithm}
\usepackage{algpseudocode}

\newcommand{\eg}{\textit{e.g.}}

\newif\ifrebuttal
\rebuttaltrue
\ifrebuttal
  
  \newcommand{\old}[1]{\textcolor{red}{\sout{#1}}}
  \newcommand{\oldlong}[1]{}
  \newenvironment{addedblock}{\par\bgroup\color{blue}}{\egroup\par}
\else
  
  \newcommand{\old}[1]{}
  \newcommand{\oldlong}[1]{}
  
\fi

\AtBeginDocument{\renewcommand{\eqref}[1]{(\ref{#1})}}

\title{Causal Scene Narration: Inference-Time Text Restructuring\\for Vision-Language-Action Driving}

\author{\name Yun Li$^{1}$ \email li-yun@g.ecc.u-tokyo.ac.jp \\
      \name Yidu Zhang$^{1}$ \\
      \name Simon Thompson$^{2}$ \\
      \name Ehsan Javanmardi$^{1}$ \\
      \name Manabu Tsukada$^{1}$ \\
      \addr $^{1}$Graduate School of Information Science and Technology, The University of Tokyo, Tokyo, Japan \\
      $^{2}$TIER IV, Inc., Tokyo, Japan}

\def\month{10}
\def\year{2026}
\def\openreview{\url{https://openreview.net/forum?id=sJATIdx6jG}}

\begin{document}

\maketitle

\begin{abstract}
Vision-Language-Action (VLA) models for autonomous driving receive navigation commands and hazard notices as disconnected text fragments, forcing the model to discover on its own which environmental constraints are relevant to the current maneuver. We introduce \textbf{Causal Scene Narration (CSN)}, which restructures VLA text inputs through intent-constraint alignment, quantitative grounding, and structured separation at inference time with zero GPU cost. To test CSN across different model states and safety layers, we evaluate it alongside training-time preference alignment (PL-DPO-NLL) and runtime supervision as experimental conditions. A multi-town CARLA evaluation (16 routes, 8 towns, $N{=}5$, 95\% bootstrap CIs) shows that CSN yields relative improvements in CARLA Driving Score (DS, route completion weighted by infraction penalty) of +31.1\% on original LMDrive and +24.5\% on the PL-DPO-NLL variant. A controlled ablation separating information content from text organization reveals that information content carries most of CSN's gain across both model weights (+6.17 DS on original LMDrive; +6.88 DS on PL-DPO-NLL). While causal structure accounts for an estimated 39.1\% (95\% CI [6.6\%, 71.4\%]) of the gain on original LMDrive versus 13.5\% on PL-DPO-NLL, the structural share on the aligned model and the cross-configuration asymmetry remain statistically inconclusive at $N{=}5$ due to wide bootstrap intervals. A perception noise ablation further confirms that CSN's benefit survives distance errors up to $\pm$5\,m, speed noise up to $\pm$30\%, and 20\% actor miss rates. Finally, we report a negative result on combining CSN with runtime supervision: a naive direction-conflict monitor never fires on this benchmark and, when paired with CSN, degrades DS via passive control clamping that truncates CSN-guided evasive maneuvers. Code is available at \url{https://github.com/LiYun0607/CausalSceneNarration}.
\end{abstract}

\section{Introduction}
\label{sec:introduction}

Vision-Language-Action (VLA) models combine visual perception with language-conditioned action prediction for end-to-end autonomous driving \citep{shao2024lmdrive, tian2024drivevlm, mao2023gpt}. In these systems, camera images and LiDAR point clouds are encoded by vision backbones, while natural language provides navigation goals and scene context. Recent results show that text input quality substantially affects driving performance. TLS-Assist \citep{schmidt2024tlsassist} improved LMDrive's driving score by 14.1\% simply by injecting structured traffic light messages, and GraphPilot \citep{schmidt2026graphpilot} achieved a 15.6\% gain through scene graph serialization, both without any retraining.

\begin{figure}[!t]
\centering
\includegraphics[width=0.8\textwidth]{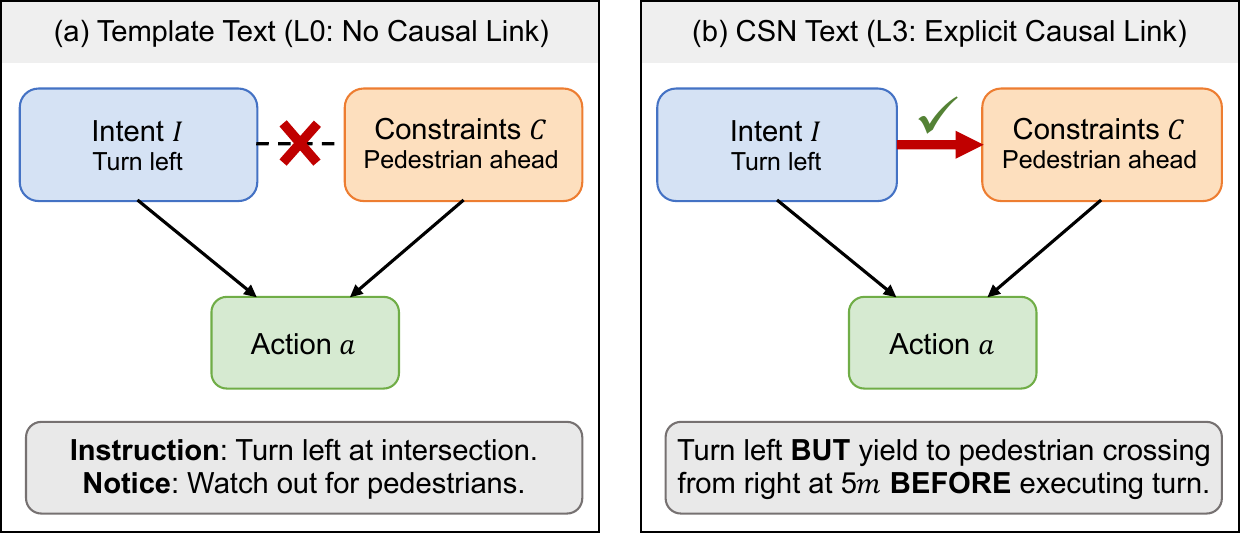}
\caption{The text structure bottleneck. (a)~Template text presents intent $\mathcal{I}$ and constraints $\mathcal{C}$ as independent fragments with no $\mathcal{I} \to \mathcal{C}$ link. (b)~CSN restores the causal link via explicit connectives (\textbf{BUT}, \textbf{BEFORE}).}
\label{fig:causal_comparison}
\end{figure}

A natural interpretation is that enriched text simply carries more scene information. While added information content indeed accounts for a large portion of these gains, we argue that a second operative variable is causal structure: whether the text explicitly links what the agent intends to do with what the environment requires it to consider (illustrated in Fig.~\ref{fig:causal_comparison}). In causal inference \citep{pearl2009causality}, observing that two variables co-vary does not tell us whether intervening on one will change the other. Similarly, presenting ``Turn left'' and ``Pedestrian ahead'' as co-occurring fragments does not tell the model whether the pedestrian is relevant to the turn. Current VLA systems face two related challenges when deploying such text interventions.

First, existing systems generate navigation commands and hazard notices as \textit{causally unrelated fragments}. LMDrive \citep{shao2024lmdrive}, for example, presents ``Turn left'' and ``Pedestrians ahead'' separately. The model must independently discover through implicit cross-attention that the pedestrian is relevant \textit{because} the left turn will cross its trajectory. A human instructor would instead say: ``Turn left, \textit{but} yield to the pedestrian at 12\,m.'' While recent frameworks such as DriveVLM \citep{tian2024drivevlm}, DriveLM \citep{sima2024drivelm}, and SteerVLA \citep{gao2026steervla} also link intent to constraints through structured schemas or question-answering graphs, such rigid formats typically require task-specific fine-tuning to parse. In contrast, encoding intent-constraint links via natural-language causal connectives allows a frozen VLA backbone to condition on causal structure without any weight modification.

Second, training-time alignment and runtime safety layers interact non-trivially with inference-time prompt interventions. Preference optimization methods such as DPO \citep{rafailov2023dpo} and Multi-PrefDrive \citep{li2025multipref} alter model weights to improve driving behavior, which may shift how the backbone utilizes text inputs, while post-hoc runtime monitors may conflict with maneuver adjustments triggered by enriched prompts. Because inference-time text enrichment leaves model weights unchanged, understanding how it composes with preference-aligned weights and runtime supervision is essential for practical deployment.

Our primary contribution targets the first limitation: \textbf{Causal Scene Narration} (CSN, \S\ref{sec:csn}) restructures VLA text inputs around intent-constraint causal alignment, quantitative physical grounding, and structured information separation. The resulting narration mirrors the perception-prediction-planning reasoning chain rather than presenting disconnected observations, and the entire pipeline runs on CPU with zero additional GPU memory. To probe how CSN composes with other intervention axes, we evaluate two additional experimental conditions rather than presenting them as standalone contributions: training-time preference alignment via Plackett--Luce DPO with NLL regularization (PL-DPO-NLL, \S\ref{sec:training}), which provides a second weight configuration for ablation, and a runtime direction-conflict monitor (\S\ref{sec:safety_supervisor}), which applies inference-time filtering on top of the VLA output.

We test the role of causal structure through a controlled empirical decomposition (\S\ref{sec:empirical_decomposition}) that separates scene information content from text organization across both weight configurations. While added scene information drives the majority of the overall improvement, causal structure accounts for a point estimate of 39.1\% (95\% bootstrap CI [6.6\%, 71.4\%], $P({>}0){=}0.986$) of CSN's gain on original LMDrive, with the lower CI bound above zero, and 13.5\% (95\% CI [$-$244\%, 41.9\%], $P({>}0){=}0.664$) on the PL-DPO-NLL variant. A bootstrap test for the cross-configuration asymmetry yields a 95\% CI [$-$23.8\%, 306.4\%] that spans zero; accordingly, we present this asymmetry as a suggestive pattern rather than an established effect.

Our contributions are:
\begin{enumerate}
    \item We identify the absence of explicit intent-constraint links as a key limitation of standard VLA text inputs, organize existing approaches into a taxonomy (L0--L3) by their level of structured linking, and propose CSN, a zero-VRAM text enrichment pipeline built on intent-constraint alignment, quantitative physical grounding, and structured information separation.
    \item An evaluation across 16 routes, 8 towns, and $N{=}5$ repetitions with 95\% bootstrap confidence intervals (CIs) shows that CSN benefits both tested weight configurations (Original LMDrive and a PL-DPO-NLL variant) with overlapping CIs. Controlled ablations show that while added scene information carries most of the gain on both configurations, the share attributable to causal phrasing is statistically measurable on Original LMDrive, indistinguishable from zero on PL-DPO-NLL, and exhibits a cross-configuration difference within statistical noise (\S\ref{sec:empirical_decomposition_results}). At the route level, the aggregate improvement is driven by large gains on high-leverage routes (5 wins, 4 ties, and 7 losses at a 0.5-DS margin on Original LMDrive) rather than a uniform lift across all scenarios.
    \item We re-implement the text-structural patterns of DriveVLM and DriveLM under a matched token budget on the same LMDrive benchmark, demonstrating that CSN's natural-language causal connectives transfer zero-shot to a frozen LMDrive backbone whereas schema-label and Q\&A-graph patterns do not (\S\ref{sec:baseline_comparison}).
    \item As a negative result on naive runtime supervision, we find that pairing a direction-conflict monitor with CSN \emph{degrades} driving performance via passive control clamping, even though the monitor's explicit conflict specifications never fire. We characterize the underlying mechanism (\S\ref{sec:component_interaction}) and discuss design implications for safety layers deployed alongside text-enriched VLA models.
\end{enumerate}

%%%%%%%%%%%%%%%%%%%%%%%%%%%%%%%%%%%%%%%%%%%%%%%%%%%%%%%%%%%%%%%%%%%%%%%%%%%%%%%%
\section{Related Work}
\label{sec:related_work}

\subsection{VLA Models and Text Structure for Driving}

End-to-end autonomous driving increasingly uses VLA models for closed-loop control. GPT-Driver \citep{mao2023gpt} reformulated motion planning as language modeling, and showed that autoregressive language models can generate plausible driving trajectories from text-encoded scene states. DriveVLM \citep{tian2024drivevlm} introduced a three-stage Chain-of-Thought (CoT) pipeline consisting of scene description, scene analysis, and hierarchical planning, and showed that when text mirrors the causal reasoning process, long-tail scenario handling improves. DriveLM \citep{sima2024drivelm} employed graph-structured visual question answering to create logical dependency chains between perception, prediction, and planning nodes, explicitly encoding causal relationships.

LMDrive \citep{shao2024lmdrive} achieves closed-loop control via Q-Former alignment of visual tokens with text, using LLaVA-v1.5 as backbone. Its template-based instruction planner generates navigation commands and hazard notices as causally disconnected fragments, which sits at the minimal end of the text structure spectrum. SteerVLA \citep{gao2026steervla} showed that replacing sparse routing commands with fine-grained meta-actions improved driving score by 4.77 points on Bench2Drive, where meta-actions carry implicit causal structure.

Among text-enrichment approaches, those that outperform LMDrive consistently provide text with richer causal structure, whether through CoT decomposition, graph-structured QA, multi-channel separation, or fine-grained meta-actions.

Several studies demonstrate that text enrichment works even without retraining. TLS-Assist \citep{schmidt2024tlsassist} achieved +14.1\% DS on LMDrive by injecting structured traffic light messages \textit{without retraining}, showing that LMDrive's pre-trained LLaMA backbone can use structured text never seen during fine-tuning. GraphPilot \citep{schmidt2026graphpilot} achieved +15.6\% through scene graph serialization, where relational structure (`pedestrian \textit{is-crossing} ego-lane \textit{conflicts-with} intended left turn') \rTwo{is consistent with} the causal structure hypothesis. SimLingo \citep{wayve2025simlingo} found no improvement from post-hoc CoT narration. A possible explanation is that their CoT narrated intended actions without connecting environmental observations to action decisions.

\subsection{Runtime Safety for Autonomous Driving}
\label{sec:related_safety}

Existing runtime safety approaches occupy two categories. \textit{Formal frameworks} include Responsibility-Sensitive Safety (RSS) \citep{shalev2017rss}, which defines safe distance envelopes and triggers proper responses (braking) when violated; Control Barrier Functions (CBFs) \citep{ames2019cbf}, which enforce forward invariance of safe sets; and Simplex switching \citep{sha2001simplex, phan2020neural_simplex}, where a verified safety controller runs alongside an unverified high-performance controller. \textit{Runtime verification} methods include STL monitoring \citep{desai2017stl_monitoring} and shield synthesis \citep{alshiekh2018shielding}, which check controller behavior against temporal logic specifications.

All these methods operate on physical state (distances, velocities) and are not designed to detect \textit{semantic-level} VLA failures such as direction misinterpretation or hallucinated scene elements. Recent end-to-end driving methods lack explicit runtime safety layers. UniAD \citep{hu2023uniad} jointly optimizes perception through planning but errors propagate unchecked. VAD \citep{jiang2023vad} introduces planning constraints that are training-time loss functions not enforced at inference. \citet{wu2022tcp} and \citet{jaeger2023hidden_biases} document that waypoint predictions fail specifically at junctions due to a ``target point shortcut'' where models steer toward the nearest GPS waypoint rather than following road geometry. \rOne{The direction-conflict monitor we evaluate as an experimental condition (\S\ref{sec:safety_supervisor}) targets one instance of this failure mode, specifically direction inconsistency during junction approach.}

\rOne{This monitor} instantiates a Simplex switch \citep{sha2001simplex} with a safety envelope reformulated for the semantic domain, targeting direction consistency and liveness rather than physical distance maintenance.

\subsection{Training-Time Safety Alignment}

DPO \citep{rafailov2023dpo} and its variants optimize policies on preference data but face probability collapse, where chosen action likelihoods decrease during optimization \citep{pang2024iterative, liu2024likelihood}. Multi-PrefDrive \citep{li2025multipref} applied multi-preference tuning to LLM-based autonomous driving and reported improved in-distribution performance through Plackett-Luce ranking \citep{plackett1975analysis} over multiple candidate actions. NLL regularization provides an explicit likelihood floor against probability collapse. We combine both in our PL-DPO-NLL objective (\S\ref{sec:training}).

%%%%%%%%%%%%%%%%%%%%%%%%%%%%%%%%%%%%%%%%%%%%%%%%%%%%%%%%%%%%%%%%%%%%%%%%%%%%%%%%
\section{Theoretical Foundations}
\label{sec:theory}

\subsection{Text Structure as a Performance Bottleneck}
\label{sec:causal_bottleneck}

\rThree{We organize text-structural approaches on a causal structure level axis: L0 corresponds to isolated commands with no environmental context; L1 to structured factual information; L2 to entity-level scene graphs or schemas; and L3 to explicit causal dependence between intent and constraints. Section~\ref{sec:taxonomy} details this taxonomy.}

In structural causal models \citep{pearl2009causality}, the presence or absence of a directed arrow between two variables encodes a causal assumption. An arrow from $X$ to $Y$ asserts that $X$ influences $Y$, while a missing arrow asserts independence. The same logic applies to VLA text inputs. Let $\mathcal{I}$ denote the navigation intent (\eg, ``turn left'') and $\mathcal{C} = \{c_1, \ldots, c_K\}$ the environmental constraints (\eg, pedestrians, vehicles, traffic lights). The correct driving action $\bm{a}$ depends not on $\mathcal{I}$ and $\mathcal{C}$ separately, but on their causal interaction, i.e., which constraints are relevant \textit{given} the current intent. A left-turn intent makes a crossing pedestrian safety-critical; the same pedestrian is irrelevant during straight driving.

Template systems present $\mathcal{I}$ and $\mathcal{C}$ as independent text fragments, with no link connecting them. Conceptually, this is equivalent to omitting the edge $\mathcal{I} \rightarrow \mathcal{C}$ from a dependency graph: the text encodes both variables but not their relationship. The LLM must recover the missing dependency internally through multi-layer cross-attention \citep{vaswani2017attention}, without any explicit signal indicating which constraints are relevant to the current intent. Prior work supports the broader claim that text acts as a reasoning scaffold. LMDrive's own ablation on the LangAuto benchmark \citep{shao2024lmdrive} showed that adding notice instructions significantly reduces collisions, even though the visual information was always available. TLS-Assist \citep{schmidt2024tlsassist} showed the same pattern: the model could always see traffic lights in the image, but without explicit textual mention, those visual features were insufficiently weighted. These results suggest that text directs the model's attention \citep{wei2022chain} rather than just adding information. Whether \textit{causal structure} within the text provides additional benefit beyond information content is the specific question our CSN~vs.~Flat Text ablation addresses (\S\ref{sec:empirical_decomposition_results}).

CSN restores the missing link by explicitly encoding which constraints are relevant \textit{given} the current intent, using linguistic causal connectives to express this conditional relationship (Fig.~\ref{fig:causal_comparison}). With $K$ constraints, the model must evaluate $O(K)$ potential pairings to discover which ones matter for the current intent, whereas CSN pre-selects the $R \ll K$ relevant ones.

\subsection{How CSN Restructures Text}
\label{sec:three_principles}

\begin{figure*}[!t]
\centering
\includegraphics[width=\textwidth]{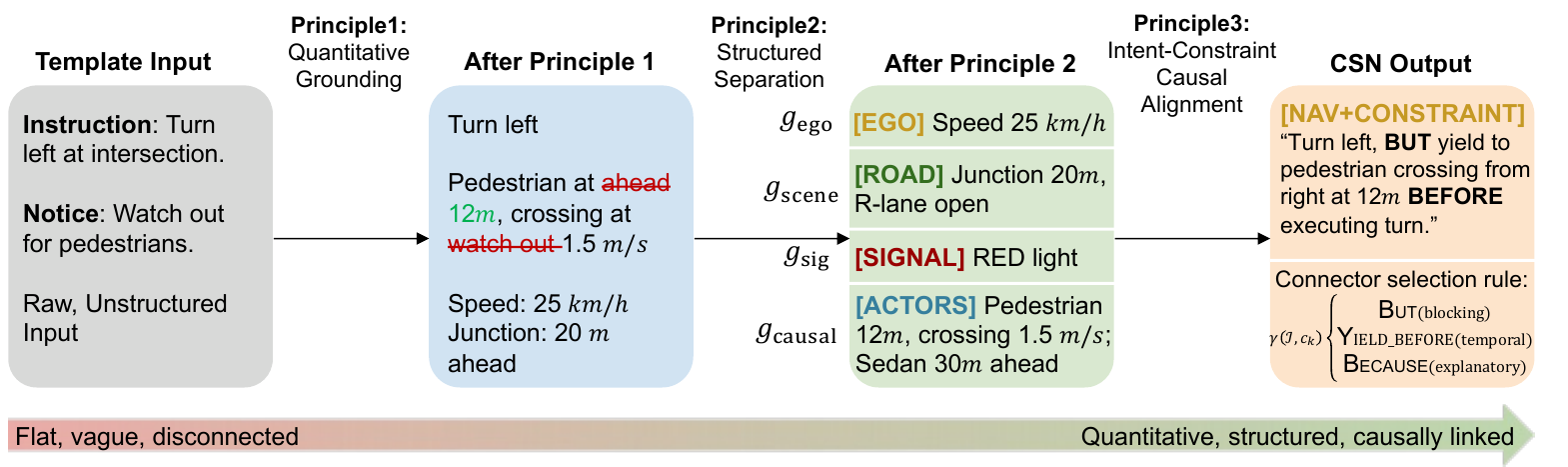}
\caption{CSN pipeline illustrated on a left-turn scenario. Each principle progressively transforms flat template text into quantitative, structured, and causally linked input.}
\label{fig:csn_pipeline}
\end{figure*}

CSN transforms standard template text through three operations, each targeting a specific weakness of flat VLA text inputs. Fig.~\ref{fig:csn_pipeline} illustrates the full pipeline on a left-turn scenario.

\textbf{(1) Quantitative physical grounding.} Template terms like `ahead' and `watch out' carry no physical dimensions. The LLM cannot tell whether a threat is 10\,m or 50\,m away, yet this difference dictates the required response \citep{shalev2017rss}. CSN replaces all vague qualifiers with exact metric values: distances in meters, speeds in km/h, and timing in seconds. For instance, translating ``watch out for pedestrians'' into ``Pedestrian at 12\,m, crossing at 1.5\,m/s'' allows the model to estimate a 4-second safety window.

\textbf{(2) Structured information separation.} Following findings that structured representation outperforms flat descriptions \citep{tian2024drivevlm, schmidt2026graphpilot}, CSN organizes the environment state into a four-part sequence (denoted $\langle \cdot \rangle$ for ordered concatenation) mirroring the perception-prediction-planning chain:
\begin{align}
    \bm{l}_{\text{CSN}} = \big\langle\; &g_{\text{ego}}(\bm{x}),\;\; g_{\text{scene}}(\mathcal{M}, \mathcal{W}),\notag\\
    &g_{\text{sig}}(\mathcal{T}),\;\; g_{\text{causal}}(\mathcal{I}, \mathcal{C}_R, \gamma) \;\big\rangle\,.
    \label{eq:structured_text}
\end{align}
Here, $g_{\text{ego}}$ encodes ego-state $\bm{x} = (v, \omega)$; $g_{\text{scene}}$ describes road topology $\mathcal{M}$ and weather $\mathcal{W}$; and $g_{\text{sig}}$ details traffic signals $\mathcal{T}$. These first three components act as independent observation encoders. The final component, $g_{\text{causal}}$, is where the reasoning occurs: it fuses the navigation intent $\mathcal{I}$ with a filtered set of relevant constraints $\mathcal{C}_R \subseteq \mathcal{C}$, selecting which detections matter for the current maneuver.

\textbf{(3) Intent-constraint causal alignment.} The third operation links intent to constraints using explicit causal connectives. Human instructors use connectives to reveal the \textit{nature} of a conflict (\eg, ``Turn left, \textsc{but}\ldots'' vs.\ ``Reduce speed \textsc{because}\ldots''). CSN mechanizes this by classifying each relevant constraint $c_k \in \mathcal{C}_R$ into one of three conflict types ($\tau_k$) and assigning a connective via a selection function $\gamma$:
\begin{equation}
    \gamma(\mathcal{I}, c_k) = \begin{cases}
        \textsc{But} & \text{if } \tau_k = \text{blocking} \\
        \textsc{Yield\_Before} & \text{if } \tau_k = \text{temporal} \\
        \textsc{Because} & \text{if } \tau_k = \text{explanatory}
    \end{cases}\,.
    \label{eq:connector}
\end{equation}

To determine $\tau_k$, let $z_k$ denote the spatial zone of the constraint $c_k$, and $\mathcal{Z}_\mathcal{I}$ denote the conflict-side zones for the ego-intent $\mathcal{I}$ (\eg, $\mathcal{Z}_{\text{left-turn}} = \{\text{ahead-left, ahead, left}\}$). The classification follows spatial-temporal rules:
\begin{itemize}
    \item \textbf{Blocking} ($\tau_k = \text{blocking}$): A stationary obstacle occupies the intended path ($z_k \in \mathcal{Z}_\mathcal{I}$). \textit{Example: a stopped vehicle directly ahead.}
    \item \textbf{Temporal} ($\tau_k = \text{temporal}$): A moving actor intersects the intended path ($z_k \in \mathcal{Z}_\mathcal{I}$). The conflict has a time dimension and may resolve if the ego yields. \textit{Example: a crossing pedestrian during a left turn.}
    \item \textbf{Explanatory} ($\tau_k = \text{explanatory}$): The constraint lies outside the immediate conflict zone ($z_k \notin \mathcal{Z}_\mathcal{I}$) but provides necessary context. \textit{Example: a speed limit reduction or a vehicle in an adjacent, non-conflicting lane.}
\end{itemize}
When a constraint satisfies multiple conditions, priority is assigned as blocking $>$ temporal $>$ explanatory. By embedding these connectives, CSN provides the LLM with explicit causal cues that it is already optimized to process from its natural language pre-training.

\rThree{The restriction to three conflict types is deliberate. Blocking, temporal, and explanatory conflicts correspond one-to-one to \textsc{but}, \textsc{yield\_before}, and \textsc{because}, connectives that LMDrive's frozen LLaMA backbone already understands from pretraining, and together they cover the four causes of route failure that dominate our 16-route benchmark. A stopped obstacle in the intended path and a red-light stop line are both blocking conflicts rendered with \textsc{but}; a crossing or merging actor is a temporal conflict rendered with \textsc{yield\_before}; a speed-limit violation is explanatory and rendered with \textsc{because}. Nothing in the $\gamma$ family of Eq.~\eqref{eq:connector} is specific to these three connectives, and further ones such as \textsc{unless} or \textsc{until} would change only the inference-time text renderer, with no retraining of the VLA. Whether three connectives remain sufficient outside driving, for example in indoor manipulation, is untested, and we record this in the limitations.}

\subsection{Empirical Decomposition of Text Utility}
\label{sec:empirical_decomposition}

CSN's design raises an empirical question: does the model benefit simply from receiving \textit{more} environmental information, or specifically from the \textit{causal organization} of that information? To answer this, we introduce an intermediate baseline $\bm{l}_{\text{disc}}$ by setting $\gamma = \varnothing$ in Eq.~\eqref{eq:structured_text}, so $g_{\text{causal}}(\mathcal{I}, \mathcal{C}_R, \varnothing)$ provides the same quantitative facts as CSN (distances, speeds, states) but presents them as disconnected fragments without causal connectives. This yields a clean three-way comparison: template $\to$ disconnected $\to$ CSN. The total driving-score gain over the baseline decomposes as:
\begin{align}
    \Delta\text{DS}_{\text{total}}
    &= \text{DS}(\bm{l}_{\text{CSN}}) - \text{DS}(\bm{l}_{\text{template}}) \notag \\
    &= \underbrace{\big[\text{DS}(\bm{l}_{\text{disc}}) - \text{DS}(\bm{l}_{\text{template}})\big]}_{\text{Utility}_{\text{info}}}
     + \underbrace{\big[\text{DS}(\bm{l}_{\text{CSN}}) - \text{DS}(\bm{l}_{\text{disc}})\big]}_{\text{Utility}_{\text{struct}}}\,.
    \label{eq:ig_decomposition}
\end{align}
Here, $\text{Utility}_{\text{info}}$ captures the gain from quantitative grounding and structured separation alone, while $\text{Utility}_{\text{struct}}$ isolates the additional gain from explicitly aligning intent with constraints. A positive $\text{Utility}_{\text{struct}}$ shows that VLA performance is bottlenecked by the model's ability to infer causal dependencies from flat text, beyond any limitation from information quantity alone. \rOne{The intermediate baseline $\bm{l}_{\text{disc}}$ is the controlled condition that allows us to measure the structural contribution; the empirical estimate is reported in \S\ref{sec:empirical_decomposition_results}.}

\subsection{Taxonomy of Text Structure Approaches}
\label{sec:taxonomy}

We organize existing VLA text-conditioning approaches by their level of \textit{causal structure} (Table~\ref{tab:taxonomy}): L0 provides isolated navigation commands with no environmental context; L1 adds structured factual observations; L2 introduces entity-level relationships, scene graphs, or fine-grained action decomposition; and L3 explicitly models the causal dependence between navigation intent and environmental constraints.

\begin{table}[!t]
\centering
\caption{Taxonomy of VLA text approaches by causal structure level. \rOne{Top block: each approach's reported gain in its original setting (not directly comparable). Bottom block: apples-to-apples results on our LMDrive benchmark, where the ``style'' rows apply each method's text-structural pattern to LMDrive at matched token budget (\S\ref{sec:baseline_comparison}).}}
\label{tab:taxonomy}
\small
\resizebox{\textwidth}{!}{%
\begin{tabular}{lccccl}
\toprule
\textbf{Approach} & \textbf{Level} & \textbf{VRAM} & \textbf{Retrain} & \textbf{DS Gain} & \textbf{Key Mechanism} \\
\midrule
LMDrive template \citep{shao2024lmdrive} & L0 & 0 & -- & baseline & Isolated instruction + notice \\
TLS-Assist \citep{schmidt2024tlsassist} & L1 & 0 & No & +14.1\% & Structured signal messages \\
GraphPilot \citep{schmidt2026graphpilot} & L2 & 0 & No & +15.6\% & Entity-relationship graph text \\
SteerVLA \citep{gao2026steervla} & L2 & 0 & Yes & +4.77 pts & Fine-grained meta-actions \\
DriveVLM CoT \citep{tian2024drivevlm} & L3 & High & Yes & -- & 3-stage causal chain \\
DriveLM graph QA \citep{sima2024drivelm} & L3 & High & Yes & -- & Graph dependency chains \\
\textbf{CSN (ours)} & \textbf{L3} & \textbf{0} & \textbf{No} & \textbf{+31.1\%} & \textbf{Intent-constraint alignment} \\
\midrule
\multicolumn{6}{l}{\textit{Apples-to-apples on the same multi-town LMDrive benchmark (\S\ref{sec:baseline_comparison})}} \\
\midrule
DriveVLM-style (compressed) & L3 & 0 & No & $-$4.7\% & Field labels (motion=/influence=) \\
DriveVLM-style (verbatim) & L3 & 0 & No & $-$74.1\% & Verbatim Critical Object Analysis \\
DriveLM-style (compressed) & L3 & 0 & No & $-$11.4\% & Q\&A graph with Context prefix \\
DriveLM-style (verbatim) & L3 & 0 & No & $-$82.7\% & Full perception--prediction--planning chain \\
\bottomrule
\end{tabular}}
\end{table}

All L3 approaches share the core property of explicitly encoding the conditional dependence between environmental constraints and navigation intent, as formalized in \S\ref{sec:causal_bottleneck}. However, among the L3 structural patterns evaluated on the same frozen LMDrive benchmark (\S\ref{sec:baseline_comparison}), CSN is the only one whose natural-language surface form aligns closely enough with the backbone's pretraining and instruction distribution to yield a positive gain over the baseline. By contrast, the schema-label and Q\&A-graph patterns adapted from DriveVLM and DriveLM, both of which presuppose end-to-end retraining on their own markup schemas, either fail to improve over the baseline under a matched token budget or severely degrade performance when rendered verbatim. To our knowledge, CSN is the first L3 framework to achieve intent-constraint causal alignment entirely at inference time with zero additional VRAM and no model retraining.

%%%%%%%%%%%%%%%%%%%%%%%%%%%%%%%%%%%%%%%%%%%%%%%%%%%%%%%%%%%%%%%%%%%%%%%%%%%%%%%%
\section{Methodology}
\label{sec:methodology}

\subsection{System Architecture Overview}
\begin{figure}[!t]
\centering
\includegraphics[width=0.95\textwidth]{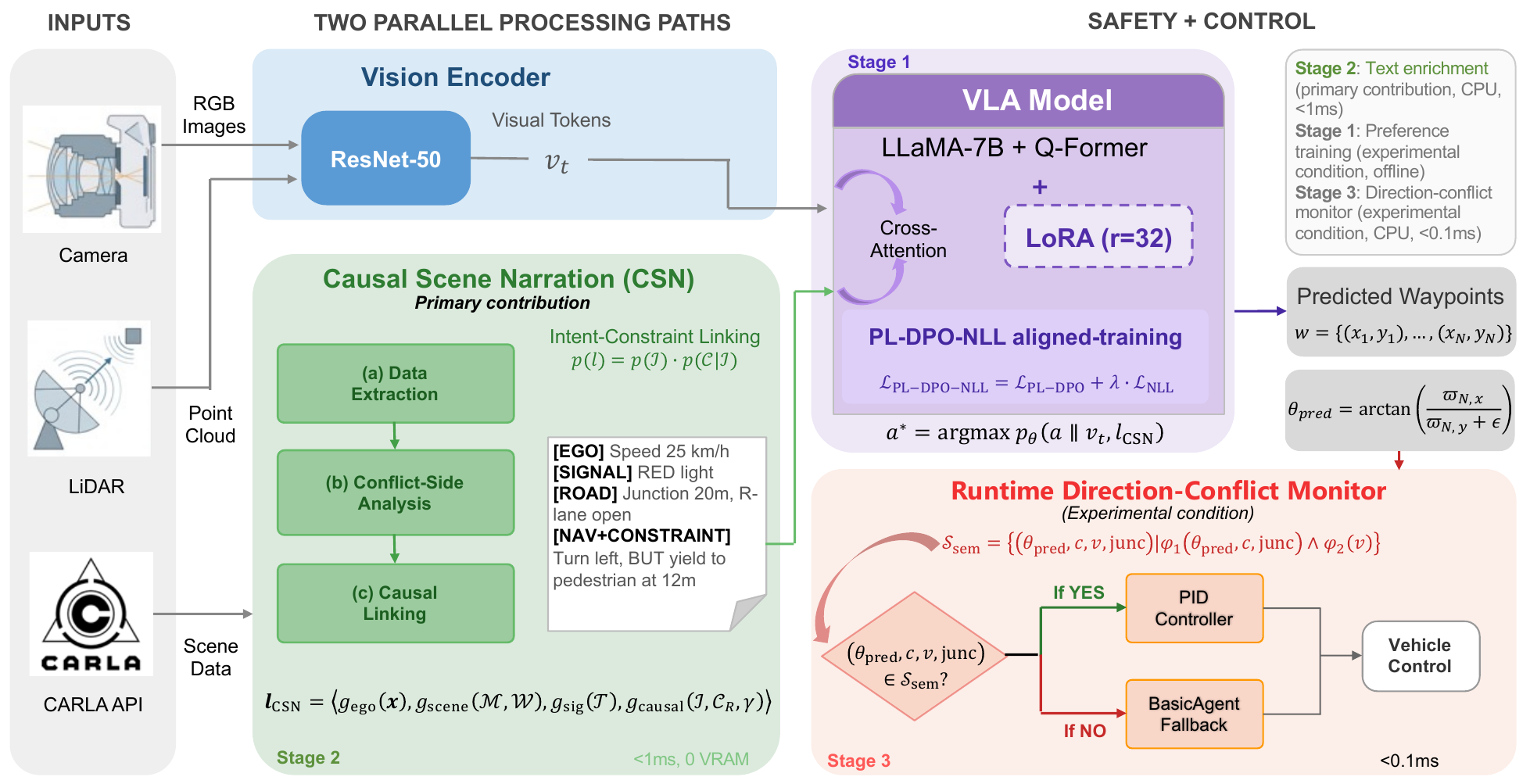}
\caption{\rOne{Architecture and pipeline. CSN (Stage~2, our primary contribution) restructures driving-environment information into structured text and runs at zero GPU cost. PL-DPO-NLL training (Stage~1) and the runtime direction-conflict monitor (Stage~3) are evaluated as experimental conditions; see \S\ref{sec:experiments}.}}
\label{fig:architecture}
\end{figure}

Our primary contribution operates at inference time (Fig.~\ref{fig:architecture}, Stage~2): Causal Scene Narration (CSN, \S\ref{sec:csn}) restructures text inputs from driving-environment data on CPU at zero GPU cost. To evaluate how CSN composes with interventions at other stages of the pipeline, we include two additional experimental conditions treated as ablations rather than standalone contributions. At Stage~1 (training time), PL-DPO-NLL (\S\ref{sec:training}) fine-tunes the base LLaMA-7B model on multi-preference rankings, providing a second weight configuration for ablation. At Stage~3 (inference time), a runtime direction-conflict monitor (\S\ref{sec:safety_supervisor}) tests whether per-frame output filtering composes with text-side enrichment.

\subsection{Causal Scene Narration Pipeline}
\label{sec:csn}
The CSN pipeline converts driving-environment observations into structured natural language following the three principles established in Section~\ref{sec:three_principles}.

\subsubsection{Environmental Data Extraction}
We extract four categories of environmental data from CARLA's Python API, all computed on CPU: (1)~dynamic actor states, including all vehicles and pedestrians within 50\,m forward and 15\,m lateral, with positions and velocities in the ego-vehicle frame and spatial zone classifications (ahead/behind/left/right; near/mid/far); (2)~traffic infrastructure, including traffic light states, elapsed phase timing, and speed limits; (3)~road topology, including junction proximity, lane availability, and curvature; and (4)~environmental conditions, including precipitation, fog density, road wetness, and sun altitude. In this work, we use CARLA's privileged API to isolate and evaluate the impact of text structure independently of upstream perception errors; we evaluate robustness under synthetic perception noise and discuss vision-based deployment in \S\ref{sec:noise_ablation}.

\subsubsection{Causal Narration Generation}
Given the navigation command $\mathcal{I}$ and detected constraints $\mathcal{C}$, the algorithm first filters $\mathcal{C}$ for relevance to $\mathcal{I}$ via conflict-side analysis, following GraphPilot \citep{schmidt2026graphpilot}. Constraints located on the conflict side of the intended maneuver (e.g., ahead-left, ahead, and left for a left turn) receive higher priority than those on non-conflict sides. The filtered constraints are then ranked by urgency based on proximity and dynamic state. Finally, each relevant constraint is linked to $\mathcal{I}$ using a causal connective selected by the $\gamma$ function defined in Eq.~\eqref{eq:connector}. The entire pipeline executes in less than 1\,ms per frame on CPU.

\begin{algorithm}[!t]
\caption{\rThree{CSN constraint filtering and ranking}}
\label{alg:csn}
\begin{algorithmic}[1]
\Require navigation intent $\mathcal{I}$, detected constraints $\mathcal{C} = \{c_1, \ldots, c_K\}$, conflict-side zones $\mathcal{Z}_\mathcal{I}$, relevance threshold $\tau$, token budget $T_{\max}$
\Ensure ordered constraint list $\mathcal{C}_R$ rendered into causal narration
\State $\mathcal{C}_{\text{kept}} \gets \varnothing$
\For{each detected constraint $c \in \mathcal{C}$ \textbf{(truncate to $\le 3$ vehicles, $\le 2$ pedestrians)}}
    \State $r \gets \textsc{Relevance}(c, \mathcal{I}, \mathcal{Z}_\mathcal{I})$ \Comment{distance-, zone-, and motion-weighted priority}
    \If{$r > \tau$}
        \State $\mathcal{C}_{\text{kept}} \gets \mathcal{C}_{\text{kept}} \cup \{(c, r)\}$
    \EndIf
\EndFor
\State Sort $\mathcal{C}_{\text{kept}}$ by $r$ in descending order
\State $\mathcal{C}_R \gets [\,]$; $\;t \gets \textsc{EstimateTokens}(\text{intent phrase})$
\For{each $(c, r) \in \mathcal{C}_{\text{kept}}$}
    \State $\tau_c \gets$ classify $c$ as blocking, temporal, or explanatory (priority order)
    \State $f \gets \textsc{Render}(c, \gamma(\mathcal{I}, c))$ \Comment{$\gamma$ defined in Eq.~\eqref{eq:connector}}
    \If{$t + \textsc{EstimateTokens}(f) + 2 > T_{\max}$} \textbf{break} \EndIf
    \State Append $f$ to $\mathcal{C}_R$; $\;t \gets t + \textsc{EstimateTokens}(f) + 1$
\EndFor
\State \Return $\mathcal{C}_R$
\end{algorithmic}
\end{algorithm}

Algorithm~\ref{alg:csn} formalizes this two-stage filter-then-rank pipeline. The \textsc{Relevance} function combines three multiplicative signals: a distance-decay term ($\max(0, 50 - d)$ for vehicles and $\max(0, 40 - d)$ for pedestrians, with distance $d$ in meters), a conflict-zone multiplier ($2.5\times$--$3.0\times$ if the constraint lies in $\mathcal{Z}_\mathcal{I}$, and $0.4\times$--$0.5\times$ otherwise), and a motion modifier ($2.0\times$ for stopped-ahead vehicles, $1.5\times$ for approaching actors, and $3.0\times$ for crossing pedestrians). We retain every constraint whose final priority score $r$ exceeds the relevance threshold $\tau = 0$. After ranking by $r$, the renderer greedily packs causal fragments until the per-channel token budget $T_{\max} = 58$ tokens is reached, which determines the effective number of surfaced constraints $K$ (variable per frame; typically $K \le 5$). In practice, the token budget, rather than a fixed $K$, is the binding constraint. Candidate vehicles and pedestrians are also capped at 3 and 2, respectively, prior to scoring as a defensive pre-filter against densely populated CARLA frames. Table~\ref{tab:text_comparison} contrasts the three text conditions used in our ablation study.

\begin{table}[!t]
\centering
\small
\caption{Text input comparison for a left-turn scenario. Causal connectors shown in bold.}
\label{tab:text_comparison}
\begin{tabular}{@{}p{0.08\textwidth}p{0.85\textwidth}@{}}
\toprule
\multicolumn{2}{@{}l}{\textbf{(a) Template (LMDrive original)}} \\
\midrule
\textit{Instr.} & Turn left at intersection. \\
\textit{Notice} & Watch out for pedestrians. \\[2pt]
\midrule
\multicolumn{2}{@{}l}{\textbf{(b) +Flat Text (same facts, no causal links)}} \\
\midrule
\textit{Instr.} & Turn left. Speed 25 km/h. Pedestrian 5m right crossing left. Sedan 12m ahead 30 km/h. RED light. Junction 20m. \\[2pt]
\midrule
\multicolumn{2}{@{}l}{\textbf{(c) CSN (causal structure)}} \\
\midrule
\textit{Instr.} & Turn left at intersection, \textbf{BUT} yield to pedestrian crossing from right at 5m \textbf{BEFORE} executing turn. Maintain distance from sedan ahead. \\
\textit{Notice} & {[EGO]} 25/30 km/h. {[ROAD]} Junction 20m, R-lane open. {[SIGNAL]} RED. {[ACTORS]} Sedan 12m ahead 30 km/h. Ped 5m R, crossing L. \\
\bottomrule
\end{tabular}
\end{table}

\subsection{Runtime Direction-Conflict Monitor (Experimental Condition)}
\label{sec:safety_supervisor}
We include a runtime monitor as an experimental condition to test whether per-frame output supervision composes with text-side enrichment. The monitor is structured around the Simplex switching architecture \citep{sha2001simplex}; we detail its formal design below and analyze its empirical interaction with CSN in \S\ref{sec:component_interaction}.

\subsubsection{Simplex Architecture for VLA Driving}
The runtime supervisor monitors whether the VLA's predicted trajectory remains inside a semantic safety envelope \citep{shalev2017rss} and triggers a controller switch when violations occur. We define the envelope as:
\begin{equation}
    \mathcal{S}_{\text{sem}} = \big\{(\theta_{\text{pred}}, c, v, \texttt{junc}) \mid \varphi_1(\theta_{\text{pred}}, c, \texttt{junc}) \wedge \varphi_2(v)\big\}\,,
    \label{eq:semantic_envelope}
\end{equation}
where $\varphi_1$ enforces direction consistency with junction-aware gating and $\varphi_2$ enforces liveness, both formalized below. We instantiate a three-component Simplex architecture \citep{sha2001simplex}, which pairs a complex High-Performance Controller (HPC) with a conservative High-Assurance Controller (HAC). In our setup, the Advanced Controller (AC, corresponding to the HPC) is the VLA model (LMDrive under either the original or PL-DPO-NLL weight configuration), which predicts ego-frame waypoint trajectories $\bm{w}_{\text{VLA}} = \{(x_i, y_i)\}_{i=1}^N$. The Baseline Controller (BC, corresponding to the HAC) is CARLA's Traffic Manager, a rule-based planner with access to the HD map and ground-truth actor states that provides a conservative fallback. The Decision Module (DM) evaluates envelope membership $(\theta_{\text{pred}}, c, v, \texttt{junc}) \in \mathcal{S}_{\text{sem}}$ at each step and switches control authority to the BC upon envelope exit.

The switching logic follows bidirectional Simplex \citep{phan2020neural_simplex}. Upon exiting $\mathcal{S}_{\text{sem}}$, control transfers from the AC to the BC for a minimum intervention period $T_{\min} = 20$ steps (approximately 1\,s at 20\,FPS). The BC retains authority until the state re-enters $\mathcal{S}_{\text{sem}}$, at which point control returns to the AC. Unlike classical kinematic safety envelopes that monitor inter-vehicle distances, our DM monitors the semantic consistency between the VLA's predicted trajectory and the active navigation command.

\subsubsection{Safety Specifications}
We define two safety properties targeting the dominant VLA failure modes identified by \citet{wu2022tcp} and \citet{jaeger2023hidden_biases}.

\textbf{Property $\varphi_1$: Direction consistency (approach phase).} When the route planner issues a turn command $c \in \{\textsc{Left}, \textsc{Right}\}$, the VLA's predicted waypoints must align with the intended turn direction. Let $\theta_{\text{pred}}$ denote the bearing angle of the final predicted waypoint $\bm{w}_N$ in the ego frame. Because predicted waypoints always lie ahead of the ego vehicle ($w_{N,y} > 0$), $\arctan$ suffices without the full $\operatorname{atan2}$ range:
\begin{equation}
    \theta_{\text{pred}} = \arctan\!\left(\frac{w_{N,x}}{w_{N,y} + \epsilon}\right)\,.
    \label{eq:bearing_angle}
\end{equation}
Expressed in Signal Temporal Logic \citep{desai2017stl_monitoring}, where $\square$ denotes ``always,'' the direction consistency specification requires:
\begin{equation}
    \varphi_1 \triangleq \square\Big(\neg\texttt{in\_junction} \wedge c = \textsc{Left} \implies \theta_{\text{pred}} < -\theta_{\text{thr}}\Big)\,,
    \label{eq:stl_direction}
\end{equation}
and symmetrically for $c = \textsc{Right}$. The threshold $\theta_{\text{thr}} = 20^\circ$ separates straight-ahead trajectories from turning trajectories and was selected empirically based on typical CARLA intersection geometries.

Property $\varphi_1$ is evaluated exclusively during the \textit{approach phase} prior to junction entry. Once inside the junction, the VLA's predicted waypoints naturally flatten in the rotating ego frame because the model predicts ``go forward'' relative to its instantaneous heading during mid-turn execution. Without junction-aware gating, this flattening induces false-positive direction conflicts (a straight maneuver is inferred while the route command remains \textsc{Left}/\textsc{Right}) and triggers unnecessary takeovers. Junction boundaries are queried directly from the CARLA HD map.

\textbf{Property $\varphi_2$: Stuck detection.} When throttle is applied, the vehicle must eventually make forward progress. Using $\diamondsuit_{[0,T]}$ to denote ``eventually within $T$ steps,'' we require:
\begin{equation}
    \varphi_2 \triangleq \square\Big(\texttt{throttle} > \tau_{\text{thr}} \implies \diamondsuit_{[0,\, T_{\text{stuck}}]}\, v > v_{\min}\Big)\,,
    \label{eq:stl_stuck}
\end{equation}
where $\tau_{\text{thr}} = 0.2$, $v_{\min} = 0.1\,$m/s, and $T_{\text{stuck}} = 30$ frames ($\approx 1.5$\,s). A typical stuck failure occurs when the VLA predicts forward waypoints into a stationary obstacle: the low-level PID controller applies throttle, but physical blockage prevents motion.

\subsubsection{Fallback Policy and Nominal Control Clamping}
Upon a $\varphi_1$ or $\varphi_2$ violation, the DM activates the BC with conservative parameters: automatic lane-change disabled, a 5\,m following distance, and a 40\% speed reduction. The BC's waypoints $\bm{w}_{\text{BC}}$ replace the VLA output for at least $T_{\min}$ steps before returning control to the AC. During active takeover, steering limits are relaxed to $1.2\times$ the nominal maximum to facilitate trajectory recovery, while throttle is capped at $0.6\times$ the nominal limit. In nominal operation (when $(\theta_{\text{pred}}, c, v, \texttt{junc}) \in \mathcal{S}_{\text{sem}}$), the supervisor wrapper passes AC waypoints to the low-level controller while applying a passive safety clamp to steering and throttle commands (detailed in \S\ref{sec:component_interaction}).

This supervisor is designed as a heuristic runtime filter rather than a formal guarantee, as its effectiveness depends on envelope parameterization. Our experiments (\S\ref{sec:component_interaction}) show that $\varphi_1$ and $\varphi_2$ never fire on this benchmark; consequently, the supervisor's empirical impact stems entirely from its passive control clamping during nominal AC operation rather than from Simplex switching. The computational overhead of the monitor is negligible ($<$0.1\,ms per step for HD map queries and bearing calculation). Because this module already instantiates a standard Simplex switch and serves strictly as an ablation condition to test compositional behavior with CSN, we do not evaluate additional external Simplex variants.

\subsection{Training-Time Safety Alignment (Experimental Condition)}
\label{sec:training}
We employ PL-DPO-NLL as a training-time experimental condition to provide a second weight configuration for ablation. The method combines Plackett-Luce multi-preference ranking \citep{plackett1975analysis} with negative log-likelihood (NLL) regularization to mitigate probability collapse during preference optimization \citep{rafailov2023dpo, pang2024iterative}.

\subsubsection{Preference Data}
We collect 51,124 Plackett-Luce preference rankings from CARLA Town01 across 67 route configurations. Each sample comprises one expert (chosen) action and 2--3 rejected actions ordered by increasing risk severity. The resulting dataset spans seven scenario categories: turns (40.2\%), normal driving (27.8\%), braking scenarios (14.7\%), speed adjustment (6.0\%), junctions (5.6\%), pedestrian interactions (3.1\%), and red-light scenarios (2.7\%).

\subsubsection{Objective Function}
The PL-DPO loss generalizes binary DPO to full rankings over $M$ candidates. Let $x = (\bm{v}_t, \bm{l})$ denote the multimodal context comprising visual features $\bm{v}_t$ and text input $\bm{l}$:
\begin{equation}
    \mathcal{L}_{\text{PL-DPO}} = -\mathbb{E}_{(x,\,y^{(1:M)})} \left[\sum_{i=1}^{M} \log \frac{\exp(\beta \cdot r_i)}{\sum_{j=i}^{M} \exp(\beta \cdot r_j)}\right]\,,
    \label{eq:pldpo}
\end{equation}
where $r_i = \log \frac{\pi_\theta(y^{(i)} \mid x)}{\pi_{\text{ref}}(y^{(i)} \mid x)}$ is the implicit reward, $y^{(1)}$ is the chosen expert action, and $y^{(2)}, \ldots, y^{(M)}$ are rejected actions ordered by increasing risk. The temperature parameter $\beta$ is scene-adaptive, assigning higher values to safety-critical scenarios to sharpen the preference distribution: $\beta = 0.35$ for turns, pedestrians, and red lights; $\beta = 0.25$ for braking; $\beta = 0.18$--$0.20$ for junctions and speed adjustment; and $\beta = 0.12$ for normal driving. These hyperparameters were selected via grid search on a held-out validation split from Town01.

The full PL-DPO-NLL objective incorporates explicit likelihood preservation on the chosen action:
\begin{equation}
    \mathcal{L}_{\text{PL-DPO-NLL}} = \mathcal{L}_{\text{PL-DPO}} + \lambda \cdot \mathcal{L}_{\text{NLL}}\,,
    \label{eq:pldponll}
\end{equation}
where $\mathcal{L}_{\text{NLL}} = -\log \pi_\theta(y^{(1)} \mid x)$ prevents the likelihood of expert actions from degrading during preference optimization. We set $\lambda = 0.1$ based on validation ablations, as larger values cause the NLL term to dominate the preference gradient.

\subsubsection{Training Configuration}
We apply LoRA adapters ($r = 32$, $\alpha = 32$) to all attention and MLP projection matrices ($q$, $k$, $v$, $o$, gate, down, up) of LLaMA-7B. Optimization uses 8-bit AdamW with a learning rate of $10^{-5}$, a per-device batch size of 4 with 8 gradient accumulation steps (effective batch size 32), 3 training epochs, a linear warmup ratio of 0.03, and BF16 mixed precision with gradient checkpointing across 3$\times$ NVIDIA RTX 6000 Ada GPUs.

%%%%%%%%%%%%%%%%%%%%%%%%%%%%%%%%%%%%%%%%%%%%%%%%%%%%%%%%%%%%%%%%%%%%%%%%%%%%%%%%
\section{Experiments}
\label{sec:experiments}

\subsection{Experimental Setup}
Our experiments are designed to answer three questions: (1)~Does CSN improve driving performance, and does this improvement hold across distinct model weight configurations? (2)~How much of CSN's gain stems from causal text structure versus added scene information? (3)~How do the two non-CSN experimental conditions, training-time preference alignment (PL-DPO-NLL) and the runtime direction-conflict monitor, compose with CSN, and does either provide complementary benefits?

We conduct closed-loop evaluations in CARLA 0.9.10 \citep{dosovitskiy2017carla} on an NVIDIA RTX 3090 Ti GPU using LMDrive with a LLaMA-7B \citep{touvron2023llama} backbone and a ResNet-50 \citep{he2016resnet} vision encoder, evaluating both the original release weights and our PL-DPO-NLL LoRA variant (\S\ref{sec:training}). The benchmark spans 16 routes across 8 towns drawn from the official CARLA Leaderboard route set, comprising 4 night routes, 5 rain routes, 3 dense fog routes, and 4 clear daytime routes. This diversity tests generalization across urban layouts, traffic densities, road topologies, and adverse weather conditions absent from single-town benchmarks. Each configuration is evaluated over $N{=}5$ independent repetitions. Throughout, we report the sample mean and 95\% bootstrap confidence interval (CI); non-overlapping CIs between two configurations indicate a statistically meaningful difference.

We evaluate ten configurations organized hierarchically (Table~\ref{tab:multitown_ablation}). On original LMDrive (without LoRA), we test: (1)~Original baseline, (2)~+CSN, (3)~+Flat Text, and (4)~+Semantic Safety. On the PL-DPO-NLL variant (with LoRA), we test: (5)~PL-DPO-NLL baseline, (6)~+TTC Safety, (7)~+Semantic Safety, (8)~+CSN, (9)~+CSN+Safety, and (10)~+Flat Text. Here, +Flat Text provides the exact same factual scene content as CSN but strips the causal connectives, enabling the empirical decomposition defined in Eq.~\eqref{eq:ig_decomposition} across both weight configurations.

\begin{figure}[!t]
\centering
\includegraphics[width=0.8\textwidth]{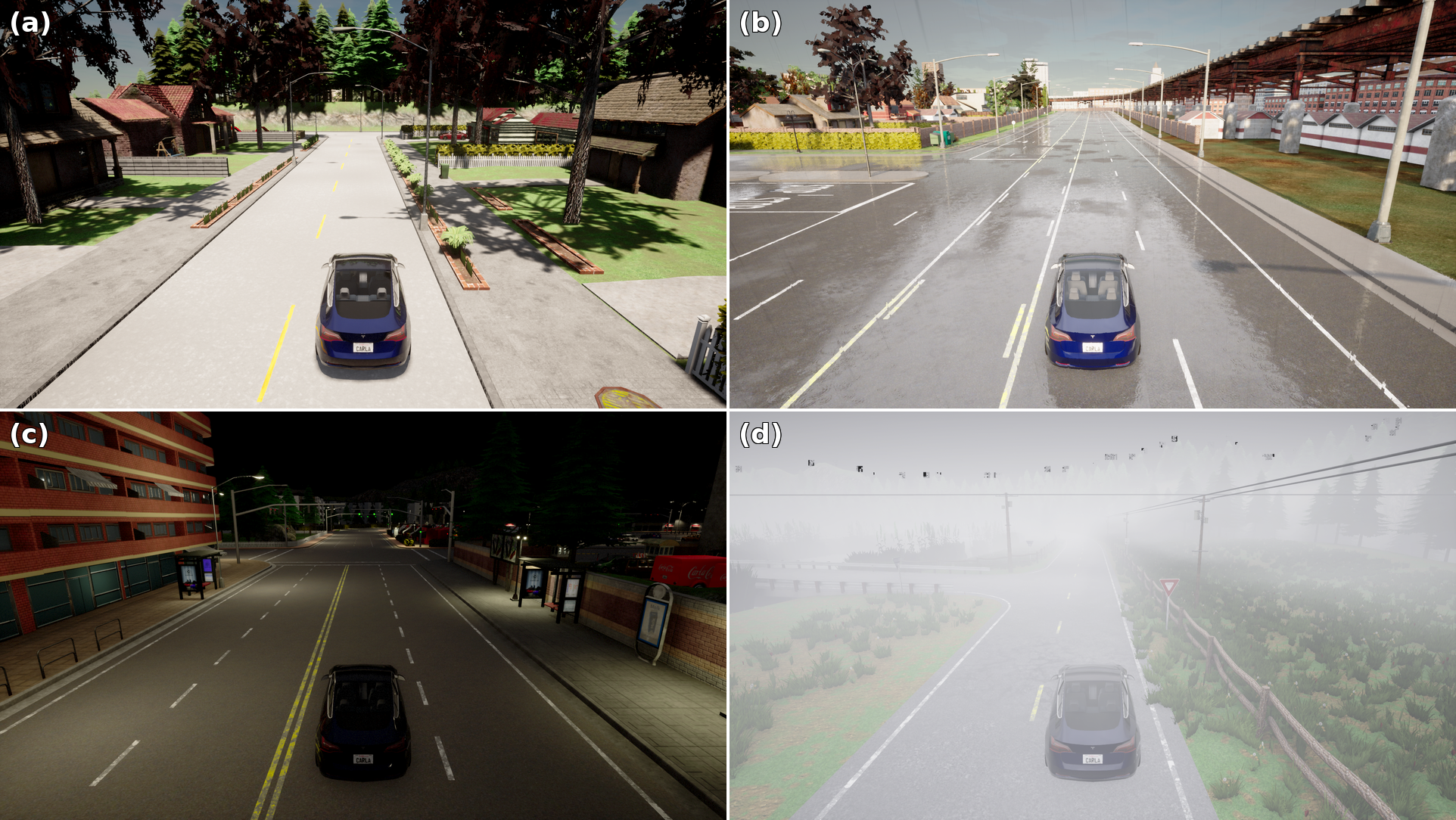}
\caption{Evaluation environments. (a)~Town01, clear day. (b)~Town03, heavy rain. (c)~Town05, night. (d)~Town07, dense fog.}
\label{fig:carla_env}
\end{figure}

We follow standard CARLA Leaderboard metrics: Driving Score~(DS), our primary metric, measures route completion weighted by an infraction penalty multiplier; Route Completion~(RC) measures the percentage of route distance completed; and Infraction Score~(IS) tracks cumulative safety penalties, where 1.0 indicates zero infractions. Fig.~\ref{fig:carla_env} illustrates representative evaluation environments.

\subsection{Main Results}
\label{sec:multitown}
Table~\ref{tab:multitown_ablation} summarizes the main results across all ten configurations along four analytical dimensions: preference training generalization, CSN robustness across weights, runtime monitor comparison, and component interaction.

\begin{table}[!t]
\centering
\caption{Multi-town ablation (16 routes, 8 towns, $N{=}5$). Values are mean $\pm$ 95\% bootstrap CI. Best \textbf{bold}, second \underline{underlined}; ties share marking.}
\label{tab:multitown_ablation}
\small
\begin{tabular}{@{}lcccrr@{}}
\toprule
\textbf{Configuration} & \textbf{DS} ($\uparrow$) & \textbf{RC} ($\uparrow$) & \textbf{IS} ($\uparrow$) & $\Delta$\textbf{DS}$_{\text{orig}}$ & $\Delta$\textbf{DS}$_{\text{base}}$ \\
\midrule
LMDrive (original) & 32.54$\pm$3.00 & 48.3$\pm$2.6\% & 0.729$\pm$0.034 & -- & -- \\
\quad + CSN & \textbf{42.67$\pm$2.74} & \textbf{56.5$\pm$1.7\%} & 0.787$\pm$0.028 & +31.1\% & -- \\
\quad + Flat Text & 38.71$\pm$1.44 & 48.7$\pm$1.9\% & \textbf{0.828$\pm$0.025} & +18.9\% & -- \\
\quad + Semantic Safety & 34.10$\pm$2.25 & 45.5$\pm$2.1\% & 0.785$\pm$0.038 & +4.8\% & -- \\
\midrule
\quad + PL-DPO-NLL & 32.49$\pm$3.34 & 44.9$\pm$2.7\% & 0.754$\pm$0.021 & $-$0.1\% & -- \\
\quad\quad + TTC Safety & 22.02$\pm$4.07 & 33.7$\pm$3.4\% & 0.658$\pm$0.037 & $-$32.3\% & $-$32.2\% \\
\quad\quad + Semantic Safety & 33.17$\pm$1.42 & 44.5$\pm$2.5\% & 0.787$\pm$0.022 & +1.9\% & +2.1\% \\
\quad\quad + CSN & \underline{40.45$\pm$3.79} & \underline{51.9$\pm$4.3\%} & 0.789$\pm$0.026 & +24.3\% & +24.5\% \\
\quad\quad + CSN+Safety & 35.74$\pm$1.37 & 48.6$\pm$2.1\% & 0.754$\pm$0.020 & +9.8\% & +10.0\% \\
\quad\quad + Flat Text & 39.38$\pm$2.66 & 49.0$\pm$1.3\% & \underline{0.823$\pm$0.047} & +21.0\% & +21.2\% \\
\bottomrule
\end{tabular}
\end{table}

\begin{figure}[!t]
\centering
\includegraphics[width=\textwidth]{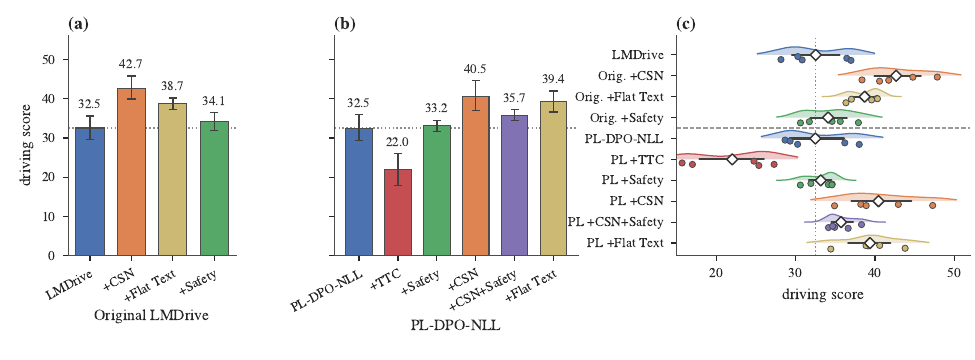}
\caption{Driving Score comparison across all ten configurations ($N{=}5$). Left and center: mean DS with 95\% bootstrap CI. Right: raincloud plot showing KDE density (shaded), individual runs (dots), mean (diamond), and 95\% CI (whisker). Dotted line: original LMDrive baseline.}
\label{fig:ds_comparison}
\end{figure}

\subsubsection{PL-DPO-NLL Does Not Improve Multi-Town Performance}
At $N{=}5$, the PL-DPO-NLL baseline and original LMDrive achieve nearly identical mean DS (32.49 vs.\ 32.54) with heavily overlapping CIs (Table~\ref{tab:multitown_ablation}), indicating that preference fine-tuning neither helps nor hurts aggregate multi-town performance. Because PL-DPO-NLL is trained on 51,124 preference rankings collected exclusively in Town01, the absence of aggregate gain across unseen towns aligns with the distribution-shift phenomenon documented in the DPO literature \citep{lin2024dpo_generalization}, where preference-aligned models overfit to training-environment dynamics at the expense of out-of-distribution generalization. Any localized improvements in Town01 are offset by regressions across the remaining seven towns. In our study, PL-DPO-NLL serves strictly as a second weight configuration to test whether CSN's overall gains and structural decomposition (\S\ref{sec:empirical_decomposition_results}) persist after preference optimization, rather than as a method claimed to improve multi-town DS. Because CSN operates exclusively on the input text without modifying model weights, it sidesteps training-induced distribution shift.

\subsubsection{CSN Benefits Hold Across the Two LoRA Weight Configurations Tested}
CSN improves mean DS by +31.1\% on original LMDrive (32.54 $\to$ 42.67) and by +24.5\% on PL-DPO-NLL (32.49 $\to$ 40.45; Table~\ref{tab:multitown_ablation}). The overlapping 95\% CIs between the two CSN-enhanced configurations ($42.67 \pm 2.74$ vs.\ $40.45 \pm 3.79$) indicate that CSN delivers comparable overall benefits across both tested weight configurations of LMDrive/LLaMA-7B. Infraction Score (IS) likewise improves on both variants, reflecting fewer traffic violations in unseen environments. We restrict this robustness claim to the two tested weight configurations and do not extrapolate to untested architectures, model scales, or training recipes.

\subsubsection{Comparing TTC and Direction-Conflict Monitors as Experimental Conditions}
The reactive time-to-collision (TTC) monitor computes kinematic TTC following \citet{hayward1972near} and triggers emergency braking whenever $\text{TTC} < 2.0$\,s, a standard threshold in the collision-warning and AEB literature \citep{vogel2003comparison}. As shown in Table~\ref{tab:multitown_ablation}, +TTC Safety yields the lowest DS and IS among all evaluated configurations. Its failure mode is systematic: frequent false-positive emergency braking causes the vehicle to repeatedly halt and restart, culminating in ``Agent got blocked'' route timeouts. Its wide CIs reflect high run-to-run variance as CARLA's stochastic traffic amplifies this over-braking pathology.

By contrast, the direction-conflict monitor (+Semantic Safety) yields narrow CIs and modest IS improvements when applied in isolation (from 0.729 to 0.785 on original LMDrive, and from 0.754 to 0.787 on PL-DPO-NLL; Table~\ref{tab:multitown_ablation}), because its conflict specifications never fire on this benchmark and thus avoid false-positive braking. However, when composed with CSN, it systematically degrades DS via passive control clamping (\S\ref{sec:component_interaction}); accordingly, we report this interaction as a negative result rather than a methodological contribution.

\subsubsection{Component Interaction Analysis}
\label{sec:component_interaction}
CSN and runtime supervision represent distinct intervention paradigms: CSN acts \textit{proactively} by enriching scene context so the VLA predicts safer trajectories, whereas the runtime monitor acts \textit{reactively} at the control output layer. Their interaction is not additive.

When layered onto the unenhanced PL-DPO-NLL baseline, the semantic supervisor yields a minor positive shift in DS ($+0.68$ DS; Table~\ref{tab:multitown_ablation}). Yet when layered on top of CSN, the exact same supervisor degrades DS by $-4.71$ points ($40.45 \to 35.74$). The tight CI of the +CSN+Safety configuration ($\pm 1.37$) confirms that this degradation is systematic rather than stochastic noise.

To isolate the underlying mechanism, we logged per-frame monitor triggers across safety-enabled runs ($N{=}3$ repetitions $\times$ 16 routes for each configuration, totaling 96 route executions across PL-DPO-NLL+Safety and +CSN+Safety). Both the direction-conflict specification ($\varphi_1$) and the stuck-detection specification ($\varphi_2$) recorded exactly zero activations across all 96 route executions. Thus, the performance drop does not stem from Simplex controller switching.

Instead, the degradation is driven by passive control clamping. The runtime monitor wrapper enforces static per-frame actuation limits ($|\text{steer}| \le 0.8$, $\text{throttle} \le 0.9$) at every timestep regardless of whether an envelope violation occurs. Without CSN, the VLA predicts conservative waypoints that translate into small steering angles, so the $|\text{steer}| \le 0.8$ bound rarely binds. With CSN, however, the model receives explicit advance warning of conflict-side hazards and initiates sharper anticipatory maneuvers, such as early lane changes or evasive steering adjustments, whose required steering commands exceed 0.8. The passive clamp truncates these evasive maneuvers mid-execution, converting valid avoidance trajectories into incomplete corrections that fare worse than unadjusted trajectories. This asymmetric binding behavior explains why the monitor slightly aids the baseline (where the clamp rarely binds) yet penalizes CSN (where the clamp truncates beneficial evasive steering). Relaxing or disabling static control clamps when CSN is active would directly resolve this conflict.

\subsection{Empirical Decomposition of Text Utility}
\label{sec:empirical_decomposition_results}
The +Flat Text ablation supplies the exact same factual scene elements as CSN while omitting causal connectives (BUT, YIELD BEFORE, BECAUSE), enabling an empirical decomposition of the total DS gain ($\Delta\text{DS}_{\text{total}}$) into information content ($\text{Utility}_{\text{info}}$) and causal structure ($\text{Utility}_{\text{struct}}$) per Eq.~\eqref{eq:ig_decomposition}. Evaluating this ablation across both weight configurations allows us to test how the decomposition behaves before and after preference tuning.

\textbf{Original LMDrive} ($N{=}5$):
\begin{align*}
    \Delta\text{DS}_{\text{total}} &= 42.67 - 32.54 = +10.13 \\
    \text{Utility}_{\text{info}} &= 38.71 - 32.54 = +6.17 \;\;(60.9\%) \\
    \text{Utility}_{\text{struct}} &= 42.67 - 38.71 = +3.96 \;\;(39.1\%)\,.
\end{align*}

\textbf{PL-DPO-NLL variant} ($N{=}5$):
\begin{align*}
    \Delta\text{DS}_{\text{total}} &= 40.45 - 32.49 = +7.96 \\
    \text{Utility}_{\text{info}} &= 39.38 - 32.49 = +6.88 \;\;(86.5\%) \\
    \text{Utility}_{\text{struct}} &= 40.45 - 39.38 = +1.08 \;\;(13.5\%)\,.
\end{align*}

\begin{figure}[!t]
\centering
\includegraphics[width=\textwidth]{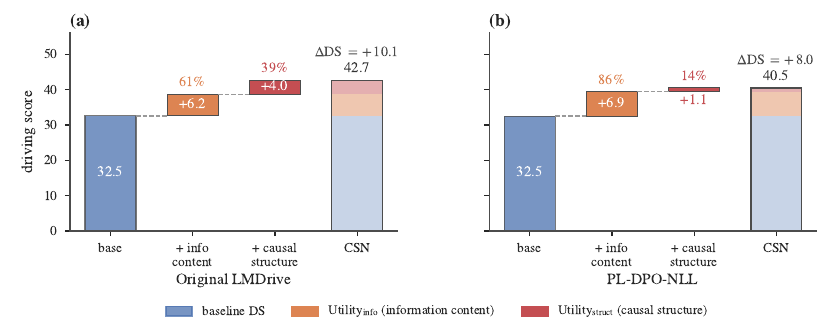}
\caption{Decomposition of CSN's DS improvement into information content and causal structure contributions on both weight configurations.}
\label{fig:ig_decomposition}
\end{figure}

On Original LMDrive, causal structure accounts for 39.1\% of the total DS gain (Fig.~\ref{fig:ig_decomposition}; 95\% bootstrap CI $[6.6\%, 71.4\%]$, $P({>}0){=}0.986$), with the lower CI bound remaining above zero. On the PL-DPO-NLL variant, the point estimate for the structural share is 13.5\% with a wide 95\% CI $[-244\%, 41.9\%]$ that spans zero ($P({>}0){=}0.664$), making it statistically indistinguishable from zero. Added scene information contributes comparably across both configurations (+6.17 DS vs.\ +6.88 DS), demonstrating that information content drives the majority of CSN's gain in both settings. At the individual route level, paired comparisons between CSN and Flat Text yield 5 wins, 4 ties, and 7 losses (0.5-DS margin) on Original LMDrive (mean difference $+3.96$ DS, 95\% CI $[-3.86, 13.41]$) and 5 wins, 6 ties, and 5 losses on PL-DPO-NLL (mean difference $+1.08$ DS, 95\% CI $[-4.44, 6.99]$), indicating that CSN's aggregate mean advantage over Flat Text is driven by substantial gains on a subset of high-leverage routes rather than a uniform lift across all 16 routes. Furthermore, a direct bootstrap test on the cross-configuration difference in structural share ($+25.6\%$) yields a 95\% CI $[-23.8\%, 306.4\%]$ that includes zero; we therefore report this cross-configuration asymmetry as a suggestive pattern rather than a statistically established effect at $N{=}5$. The observed pattern is consistent with the hypothesis that preference optimization over 51k ranked driving scenarios partially internalizes causal conflict resolution into the model weights, reducing the marginal utility of explicit textual connectives.

Finally, on both weight configurations, +Flat Text significantly outperforms its respective baseline with non-overlapping 95\% CIs (maintaining a 1.73 DS separation between CI bounds on Original LMDrive and a narrower 0.90 DS separation between CI bounds on PL-DPO-NLL), confirming that enriched scene information improves driving performance regardless of weight configuration.

\subsection{Apples-to-Apples Comparison with DriveVLM and DriveLM Text Patterns}
\label{sec:baseline_comparison}
The two existing L3 frameworks in our taxonomy (Table~\ref{tab:taxonomy}), DriveVLM \citep{tian2024drivevlm} and DriveLM \citep{sima2024drivelm}, train end-to-end VLMs jointly with their own text formats and evaluate on distinct benchmarks (nuScenes, SUP-AD, and DriveLM-CARLA). To conduct a controlled, apples-to-apples comparison on our multi-town CARLA benchmark, we extract their core text-structural patterns and inject them into LMDrive's text channels while holding the underlying VLA backbone, extracted scene state, route set, and simulation conditions strictly fixed. We intentionally avoid retraining LMDrive on these formats, as doing so would entangle the intrinsic zero-shot transferability of the text representation with weight adaptation. Holding the VLA frozen isolates the effect of text surface structure.

We implement each baseline in two variants:
\begin{itemize}
    \item \textbf{Compressed variant:} Constrained to fit LMDrive's 64-token-per-channel limit (58 tokens available after reserving the navigation intent phrase; cf.\ \S\ref{sec:csn}), matching the per-channel token budget of CSN and Flat Text.
    \item \textbf{Verbatim (faithful) variant:} Relaxes length constraints to faithfully reproduce each paper's literal layout, namely DriveVLM's three-line \texttt{Critical Object / Characteristic / Influence} schema from Fig.~1 of \citet{tian2024drivevlm} expanded with $C_s/C_m/C_b$ attribute prose, and DriveLM's full perception-prediction-planning \texttt{Q$_n$} chain with \texttt{(Context: $\cdot$)} prefixes from \S3.1 of \citet{sima2024drivelm}. Evaluating both compressed and verbatim variants tests whether token truncation alone explains any performance difference.
\end{itemize}

Table~\ref{tab:baseline_comparison} reports results on Original LMDrive ($N{=}5$ repetitions per configuration). All non-baseline rows share identical upstream scene extraction, conflict-side filtering, and urgency ranking; the sole experimental variable is the rendered text format.

\begin{table}[!t]
\centering
\caption{\rOne{Apples-to-apples comparison on Original LMDrive (16 routes, 8 towns, $N{=}5$, 95\% bootstrap CIs). All methods share identical scene state and route set; only the text format differs. Compressed variants match CSN's token budget; verbatim variants reproduce paper layouts exactly.}}
\label{tab:baseline_comparison}
\small
\begin{tabular}{@{}lcccc@{}}
\toprule
\textbf{Configuration} & \textbf{DS} ($\uparrow$) & \textbf{RC} ($\uparrow$) & \textbf{IS} ($\uparrow$) & $\Delta$\textbf{DS} \\
\midrule
LMDrive (original) & 32.54$\pm$3.00 & 48.3\% & 0.729 & -- \\
\midrule
\multicolumn{5}{l}{\textit{DriveVLM-style (3-stage CoT, motion=/influence= fields)}} \\
\quad compressed (matched budget) & 31.02$\pm$4.95 & 47.3\% & 0.672 & $-$1.5 \\
\quad verbatim Critical Object Analysis & 8.43$\pm$0.79 & 18.4\% & 0.472 & $-$24.1 \\
\midrule
\multicolumn{5}{l}{\textit{DriveLM-style (Graph VQA, Q$_n$ + Context: prefix)}} \\
\quad compressed (matched budget) & 28.82$\pm$3.30 & 47.8\% & 0.632 & $-$3.7 \\
\quad verbatim P1$\to$P2$\to$P3 chain & 5.63$\pm$0.42 & 13.7\% & 0.504 & $-$26.9 \\
\midrule
+ Flat Text (ours, no causal links) & 38.71$\pm$1.44 & 48.7\% & 0.828 & +6.2 \\
\textbf{+ CSN (ours)} & \textbf{42.67$\pm$2.74} & \textbf{56.5\%} & 0.787 & \textbf{+10.1} \\
\bottomrule
\end{tabular}
\end{table}

\textbf{Compressed variants match the unenhanced baseline.} The compressed DriveVLM-style and DriveLM-style formats achieve $31.02 \pm 4.95$ DS and $28.82 \pm 3.30$ DS, respectively, with 95\% bootstrap CIs overlapping the Original LMDrive baseline ($32.54 \pm 3.00$ DS). Substituting standard template instructions with DriveVLM's key-value \texttt{motion=/influence=} fields or DriveLM's \texttt{Q$_n$/A$_n$/Context:} chains neither improves nor degrades performance under a matched token budget. In contrast, CSN separates from the baseline cluster with a non-overlapping CI ($42.67 \pm 2.74$ DS).

\textbf{Verbatim variants severely degrade performance.} Reproducing the full schema layouts collapses Driving Score by 24--27 points below the unenhanced baseline, with narrow CIs ($\pm 0.42$ for DriveLM-faithful; $\pm 0.79$ for DriveVLM-faithful) indicating systematic failure rather than random variance. Because the verbatim configurations consume more total tokens per frame across channels than CSN (95.1 tokens for DriveLM-faithful), this collapse cannot be attributed to information loss. Frame-level inspection reveals that formal schema markers (\texttt{Object:}, \texttt{Characteristic:}, \texttt{Influence:}, \texttt{Q$_n$ (Context:~A$_k$)}) push the input out-of-distribution for LMDrive's frozen LLaMA adapter, which was fine-tuned on natural template commands rather than structured markup. Conversely, CSN's natural-language connectives (``but'', ``before'', ``because'') lie within ordinary pretraining prose that the LLM already parses fluently.

\textbf{Implications for the L3 taxonomy.} While Table~\ref{tab:taxonomy} classifies DriveVLM CoT, DriveLM graph QA, and CSN together at level L3 because all three explicitly link navigation intent to environmental constraints, our ablation reveals a crucial sub-distinction: the surface form used to encode that link. Schema-field encodings (DriveVLM) and Q\&A-graph encodings (DriveLM) are effective when models are explicitly fine-tuned on those schemas, but fail to transfer zero-shot to an unmodified VLA. Natural-language causal connectives transfer zero-shot because they align with the backbone's pretraining distribution. Thus, CSN's effectiveness stems not merely from providing L3 causal structure, but from expressing it in a pretraining-compatible surface form.

\subsection{Discussion}

\subsubsection{Why Does Causal Structure Help?}
Why does reorganizing the same facts into structured sentences improve driving performance? Consider a speed reduction scenario. Template text presents ``Reduce speed'' and ``Wet road'' as unrelated fragments, leaving the LLM to infer whether the road condition is relevant to the maneuver. CSN instead writes ``Reduce speed BECAUSE wet road reduces braking effectiveness at current 45\,km/h,'' where the connective ``BECAUSE'' serves as a direct attention cue. Because LLaMA has encountered millions of such causal constructions during natural-language pretraining, it can use them at inference time. Structured narration thus offloads part of the relational reasoning from the model's weights to the input representation.

\subsubsection{Privileged Information and Deployment Considerations}
\label{sec:noise_ablation}
As noted in \S\ref{sec:csn}, CSN currently operates on privileged simulation data. To test whether its benefits survive under realistic perception errors, we inject calibrated noise into CSN's inputs: Gaussian noise on distance measurements ($\sigma \in \{1, 2, 5\}$\,m), multiplicative noise on speed readings ($\pm\{10, 20, 30\}\%$), and random actor miss rates (0--20\%). Table~\ref{tab:noise_ablation} summarizes the results.

\begin{table}[!t]
\centering
\caption{Perception noise ablation on Original LMDrive + CSN (clean $N{=}6$, each noise condition $N{=}5$; same 16 routes and noise seed throughout). $\pm$ is the per-repetition SD and brackets give the 95\% CI, whereas Tables~\ref{tab:multitown_ablation} and~\ref{tab:baseline_comparison} report bootstrap CI half-widths. Last row: the same Severe noise applied to Flat Text as a control.}
\label{tab:noise_ablation}
\small
\begin{tabular}{lcccccc}
\toprule
\textbf{Noise Level} & \textbf{Dist.\,$\sigma$} & \textbf{Speed} & \textbf{Miss Rate} & $N$ & \textbf{DS} & \textbf{95\% CI} \\
\midrule
Clean (privileged) & 0\,m & 0\% & 0\% & 6 & $42.3 \pm 3.5$ & $[39.5,\; 45.0]$ \\
Mild & $\pm$1\,m & $\pm$10\% & 0\% & 5 & $45.2 \pm 2.9$ & $[42.6,\; 47.8]$ \\
Moderate & $\pm$2\,m & $\pm$20\% & 0\% & 5 & $45.4 \pm 2.7$ & $[43.1,\; 47.6]$ \\
Severe & $\pm$5\,m & $\pm$20\% & 10\% & 5 & $46.0 \pm 4.1$ & $[42.5,\; 49.6]$ \\
Extreme & $\pm$5\,m & $\pm$30\% & 20\% & 5 & $45.1 \pm 2.7$ & $[42.7,\; 47.4]$ \\
\midrule
Flat Text + Severe (control) & $\pm$5\,m & $\pm$20\% & 10\% & 5 & $37.7 \pm 2.4$ & $[35.6,\; 39.7]$ \\
\bottomrule
\end{tabular}
\end{table}

All four noise conditions sit above the noise-free CSN mean ($42.3$ DS, 95\% CI $[39.5, 45.0]$, $N{=}6$). Under independence and a symmetric null (each condition equally likely to fall above or below the clean CSN baseline), the joint event ``4-of-4 above clean'' has probability $\le 1/16$. Because the four noise conditions share the same 16 routes, a fixed noise RNG seed (42) across repetitions and noise levels, and the default CARLA traffic-manager seed, the true joint probability under dependence is somewhat larger than $1/16$; nevertheless, the consistent all-above-clean pattern merits a mechanism-level explanation rather than dismissal as sampling noise.

We propose three falsifiable mechanism hypotheses, each tied to a predicted outcome under a specific control:
\begin{itemize}
    \item \textbf{(i) Regularization of the text manifold.} Small perturbations of the numeric values rendered into CSN text may push the resulting strings closer to LLaMA's pretraining distribution (e.g., a rounded ``12\,m'' measurement is more frequent in pretraining corpora than ``11.83\,m''). \emph{Falsifier:} the all-above-clean pattern should vanish when noise is applied solely to the textual rendering layer while leaving the underlying perception triggers unchanged.
    \item \textbf{(ii) Asymmetric-cost safety buffer.} Even with zero-mean Gaussian distance noise (clipped at a 1\,m floor), underestimating the gap to an obstacle is operationally safer than overestimating it: CSN encodes any small remaining gap as an earlier \textsc{yield\_before} constraint, so negative distance perturbations induce conservative yielding while positive ones carry a smaller penalty in this regime. \emph{Falsifier:} the performance lift should reverse (degrading below clean CSN) under positive-only noise injection that systematically inflates reported distances, or shift predictably under sign-controlled perturbations.
    \item \textbf{(iii) Degenerate-tie breaking.} Jitter in the numeric tokens perturbs LLM decoding away from local plateaus, breaking ties among candidate actions across repetitions. \emph{Falsifier:} because pure baseline LMDrive receives no dynamic numeric text for the noise to act upon, the appropriate control is Flat Text injected with identical noise; any text-generic tie-breaking lift should reproduce there.
\end{itemize}

There is no monotone trend across noise magnitudes: Severe ($46.0$ DS) and Extreme ($45.1$ DS) differ by less than bootstrap noise, so we do not claim a saturating fit. The noise sweep alone neither rules out nor confirms hypotheses (i)--(iii).

However, because (i) and (iii) are text-generic whereas (ii) relies on CSN's causal threshold rendering, the Flat Text control specified in (iii) discriminates between them. Under the Severe noise setting ($N{=}5$, identical noise parameters and RNG seed; final row of Table~\ref{tab:noise_ablation}), noisy Flat Text scores $37.66 \pm 2.36$ DS versus $38.71 \pm 1.88$ DS for clean Flat Text (reporting repetition-level SD here, corresponding to the $38.71 \pm 1.44$ bootstrap 95\% CI half-width in Table~\ref{tab:multitown_ablation}). The resulting difference of $-1.05$ DS (bootstrap 95\% CI $[-3.35, +1.38]$, $P(\text{diff} > 0) = 0.19$) shows no performance lift, in contrast to the $+3.7$ DS lift observed under CSN ($42.3 \to 46.0$ DS). Given the wide $N{=}5$ intervals and shared noise seed, we interpret this result as disfavoring (i) and (iii) rather than definitively confirming (ii). Directly confirming (ii) would require sign-controlled noise ablations with $N \ge 20$ runs per condition; due to the computational cost of closed-loop CARLA evaluations (which limited us to $N{=}6$ for clean CSN and $N{=}5$ elsewhere), we leave this to future work.

\subsubsection{Why LMDrive as the Evaluation Platform}
Our experiments use a single VLA architecture (LMDrive with LLaMA-7B). LMDrive is currently the only open-source VLA system that simultaneously satisfies three requirements for our study: (1)~it accepts free-form text inputs that CSN can dynamically restructure, (2)~it supports closed-loop CARLA evaluation under the standard Leaderboard protocol, and (3)~its training pipeline is publicly available, enabling our controlled PL-DPO-NLL ablation. Other prominent VLA frameworks such as DriveVLM \citep{tian2024drivevlm} and Bench2Drive \citep{jia2024bench2drive} either lack open-source training code, rely on fixed command tokens rather than free-form text, or use open-loop/proprietary evaluation setups that preclude controlled comparison. CSN's mechanism, restructuring text with causal connectives that LLMs already parse from pretraining, is architecture-agnostic in principle, but validating this generality awaits additional open-source VLA platforms with closed-loop evaluation capabilities.

\subsubsection{Robustness Across Weight Configurations}
As established in \S\ref{sec:multitown}, the overlapping CIs between the original and preference-aligned CSN configurations indicate that CSN's overall benefit holds across both tested LoRA weight configurations. However, the empirical decomposition reveals an asymmetry: causal structure accounts for 39.1\% (95\% CI [6.6\%, 71.4\%]) of CSN's gain on original LMDrive compared to 13.5\% (95\% CI [$-244\%$, $+41.9\%$]) on PL-DPO-NLL, though a direct bootstrap test on their difference ($+25.6\%$, 95\% CI [$-23.8\%$, $+306.4\%$]) spans zero. Specifically, on PL-DPO-NLL, the CIs for CSN versus Flat Text overlap substantially ($40.45 \pm 3.79$ vs.\ $39.38 \pm 2.66$), rendering the 13.5\% structural share statistically indistinguishable from zero. On original LMDrive, the overlap is marginal ($42.67 \pm 2.74$ vs.\ $38.71 \pm 1.44$), supporting a meaningful structural contribution in that setting. This pattern is consistent with, but does not statistically establish, preference learning partially internalizing the causal reasoning that explicit text structure otherwise provides. Consequently, CSN should remain compatible with other LMDrive-family checkpoints without further weight updates, though the relative balance between informational and structural gains may shift. Whether this behavior transfers to distinct architectures such as DriveVLM remains untested.

\subsubsection{Benchmark Difficulty and Weather Effects}
Our evaluation benchmark spans 4 night routes, 5 rain routes, and 3 dense fog routes alongside clear daytime scenarios. Consequently, baseline DS values (32.54 for original LMDrive and 32.49 for PL-DPO-NLL) are substantially lower than the 50--60 DS typical of single-town, clear-weather evaluations. CSN's +31.1\% relative improvement holds under these challenging multi-weather conditions.

%%%%%%%%%%%%%%%%%%%%%%%%%%%%%%%%%%%%%%%%%%%%%%%%%%%%%%%%%%%%%%%%%%%%%%%%%%%%%%%%
\section{Conclusion}
\label{sec:conclusion}

We introduced Causal Scene Narration (CSN), a framework that restructures VLA text inputs around intent-constraint causal alignment at zero GPU cost. Through a multi-town CARLA evaluation across 16 routes and 8 towns, using $N{=}5$ independent repetitions ($N{=}6$ for the clean noise-ablation baseline) and reporting 95\% bootstrap confidence intervals, we establish four main findings.

First, CSN improves Driving Score on both original LMDrive (+31.1\%) and the preference-aligned PL-DPO-NLL variant (+24.5\%), with overlapping CIs demonstrating robustness across the two tested LoRA weight configurations of LMDrive/LLaMA-7B. Second, a controlled ablation shows that added scene information carries the majority of CSN's gain on both weight configurations (+6.17 DS on Original LMDrive; +6.88 DS on PL-DPO-NLL); the share attributable to causal structure is 39.1\% (95\% bootstrap CI $[6.6\%, 71.4\%]$) on Original LMDrive and 13.5\% (95\% CI $[-244\%, 41.9\%]$) on PL-DPO-NLL, with a cross-configuration difference (95\% CI $[-23.8\%, 306.4\%]$) that spans zero. This asymmetry is consistent with, though does not statistically establish, preference learning partially internalizing causal reasoning. Third, an apples-to-apples comparison against DriveVLM-style and DriveLM-style text patterns under a matched token budget (\S\ref{sec:baseline_comparison}) demonstrates that CSN's natural-language causal connectives transfer zero-shot to a frozen LMDrive backbone whereas schema-label and Q\&A-graph patterns do not, isolating the benefit of CSN's surface form within the L3 taxonomy class. Fourth, a perception noise ablation confirms that CSN's benefit survives distance errors up to $\pm$5\,m, speed noise up to $\pm$30\%, and 20\% actor miss rates, indicating that the gains do not rely on numerically exact perception inputs.

We additionally report a negative result on layering a runtime direction-conflict monitor on top of CSN. Although the monitor's conflict specifications never fire on this benchmark, its passive control clamping truncates CSN-guided evasive steering and degrades overall driving performance. Relaxing or adapting control clamps when CSN is active would mitigate this conflict, highlighting an important design consideration for runtime supervisors paired with text-enriched VLA models.

\textbf{Limitations.} (1)~Our evaluation is simulation-based; while noise injection experiments (\S\ref{sec:noise_ablation}) show robustness to synthetic sensor errors, real-world deployment requires integration with a physical perception stack, and a sim-to-real gap remains because CSN inputs are derived from CARLA's privileged API. While the noise ablation provides initial evidence of error tolerance, it does not substitute for closed-loop on-vehicle validation. (2)~Experiments evaluate a single backbone architecture and scale (LMDrive with LLaMA-7B) across two LoRA weight configurations. (3)~The runtime supervisor evaluated as an experimental condition employs fixed control clamps that do not adapt to CSN-enriched context quality, driving the negative result documented above. (4)~The $\gamma$ connective set is restricted to three causal connectives (\textsc{but}, \textsc{yield\_before}, \textsc{because}), which suffices for the conflict structure of this driving benchmark but remains untested in broader domains. (5)~Due to the computational cost of closed-loop CARLA evaluations, our perception-noise experiments were limited to $N{=}6$ for the clean CSN condition and $N{=}5$ for noisy variants (whereas $N \ge 20$ would be required to resolve the all-above-clean pattern with high statistical power).

\textbf{Future Work.} (1)~Integrating CSN with an onboard vision-based perception pipeline for real-world closed-loop validation. (2)~Extending CSN to additional VLA architectures and foundation model scales. (3)~Designing context-aware runtime supervisors whose control clamps adapt dynamically to text-enriched VLA maneuvers to avoid the truncation effect identified in \S\ref{sec:component_interaction}. (4)~Extending the empirical decomposition to token-level VLA architectures where action distributions are directly tractable, enabling a strict information-theoretic evaluation of $\text{Utility}_{\text{struct}}$.

\paragraph{Ethics Statement.} CSN aims to improve autonomous driving safety in VLA models through structured text conditioning. All evaluations are conducted exclusively in the open-source CARLA simulator with no human subjects involved; real-world deployment would require rigorous physical safety validation.

\paragraph{Reproducibility Statement.} All experiments use the publicly available CARLA 0.9.10 simulator and the open-source LMDrive codebase. Evaluation follows the standard CARLA Leaderboard protocol across 16 routes in 8 towns. Each configuration is evaluated over $N{=}5$ independent repetitions ($N{=}6$ for the clean noise-ablation baseline) with fixed random seeds, and 95\% bootstrap CIs are reported throughout. The CSN text generation pipeline, runtime monitor implementation, evaluation scripts, and trained LoRA weights are publicly available at \url{https://github.com/LiYun0607/CausalSceneNarration}.

%%%%%%%%%%%%%%%%%%%%%%%%%%%%%%%%%%%%%%%%%%%%%%%%%%%%%%%%%%%%%%%%%%%%%%%%%%%%%%%%
\bibliographystyle{tmlr}
\subsubsection*{Acknowledgments}
This work was supported by JSPS KAKENHI Grant Number JP26KJ0799 and by TIER IV, Inc.

\bibliography{references}

\end{document}